\documentclass[letterpaper,10pt,conference]{IEEEtran}
\usepackage{amsfonts,amsmath}
\usepackage{graphicx}
\usepackage{diagbox}
\usepackage{subfigure}
\usepackage{hyperref}
\IEEEoverridecommandlockouts
\makeatletter
\def\ps@IEEEtitlepagestyle{
  \def\@oddfoot{\mycopyrightnotice}
  \def\@evenfoot{}
}
\def\mycopyrightnotice{
  {\footnotesize
  \begin{minipage}{\textwidth}
  \centering
  \copyright 2021 IEEE. Personal use of this material is permitted.  Permission from IEEE must be obtained for all other uses, in any current or future media, including reprinting/republishing this material for advertising or promotional purposes, creating new collective works, for resale or redistribution to servers or lists, or reuse of any copyrighted component of this work in other works. DOI: \href{https://ieeexplore.ieee.org/document/9565000}{10.1109/ITSC48978.2021.9565000}
  \end{minipage}
  }
}

\def\cA{{\cal A}}

\def\cO{{\cal O}}

\def\cS{{\cal S}}

\def\bbE{{\mathbb E}}

\def\bbN{{\mathbb N}}

\def\bbR{{\mathbb R}}

\begin{document}
\title{\LARGE \bf
Reinforcement Learning for Mixed Autonomy Intersections
}

\author{Zhongxia Yan$^{1}$ and Cathy Wu$^{2}$

\thanks{$^{1}$Electrical Engineering and Computer Science, Massachusetts Institute of Technology {\tt\small zxyan@mit.edu}}%

\thanks{$^{2}$Laboratory for Information \& Decision Systems, Massachusetts Institute of Technology {\tt\small cathywu@mit.edu}}%
}

\maketitle
\pagestyle{plain}

\begin{abstract}
We propose a model-free reinforcement learning method for controlling mixed autonomy traffic in simulated traffic networks with through-traffic-only two-way and four-way intersections. Our method utilizes multi-agent policy decomposition which allows decentralized control based on local observations for an arbitrary number of controlled vehicles. We demonstrate that, even without reward shaping, reinforcement learning learns to coordinate the vehicles to exhibit traffic signal-like behaviors, achieving near-optimal throughput with 33-50\% controlled vehicles. With the help of multi-task learning and transfer learning, we show that this behavior generalizes across inflow rates and size of the traffic network. Our code, models, and videos of results are available at \href{https://github.com/ZhongxiaYan/mixed_autonomy_intersections}{https://github.com/ZhongxiaYan/mixed\_autonomy\_intersections}.
\end{abstract}


\section{Introduction}
Recent adoption of autonomous vehicles (AVs) and 5G communication technology raise the possibility of algorithmically controlling AVs for objectives such as reducing congestion, improving fuel consumption, and ensuring safety. However, in the near future, partial adoption of AVs results in \textit{mixed autonomy} traffic. Several recent works have demonstrated that optimized control of AVs in simulated mixed autonomy contexts may significantly reduce congestion in closed traffic networks \cite{wu2017emergent} or highway scenarios \cite{vinitsky2018lagrangian}. We similarly view mixed autonomy from a transportation science perspective: developing better understanding of mixed autonomy has important implications for policy and business decisions concerning the adoption of AVs.

In this study, we demonstrate near-optimal control of AVs in unsignalized mixed autonomy intersection scenarios with model-free multi-agent reinforcement learning (RL). Specifically, we consider two-way and four-way intersection scenarios with no turn traffic. We quantify the effect of optimized AVs in microscopic simulation, where controlled vehicles (hereby referred to as AVs) comprise a fraction of all vehicles while the remaining are simulated by a car-following model of human driving.
We show that our multi-agent policy gradient algorithm learns effective policies in the mixed autonomy context, measured by comparisons with signal-based and priority-based baselines.

The purpose of this study is not to introduce mixed autonomy as an alternative for traffic signal control, but rather assess as a stepping stone for RL-based analysis of joint vehicle and traffic signal control.
Although there are numerous works studying RL in the context of traffic signal control, to the best of our knowledge, RL in the context of intersection vehicular control has not received adequate attention.
We anticipate that integration of both control mechanisms may facilitate joint optimization of throughput, safety, and fuel objectives.

Our main contributions in this study are:
\begin{enumerate}
    \item We demonstrate that mixed autonomy traffic can achieve near-optimal throughput (as measured by outflow) in unsignalized intersection networks.
    \item We characterize the performance trade-off with AV penetration rate.
    \item Without using reward shaping, we show that our policies exhibit traffic-signal-like coordination behavior.
    \item We formulate policies compatible with multiple traffic network topologies and sizes, yielding good performance in larger networks unseen at training time.
\end{enumerate}
Code, models, and videos of results are available on \href{https://github.com/ZhongxiaYan/mixed_autonomy_intersections}{Github}.

\begin{figure*}[ht]
    \centering
    \subfigure[Two-way 2x1]{\includegraphics[width=1.36in]{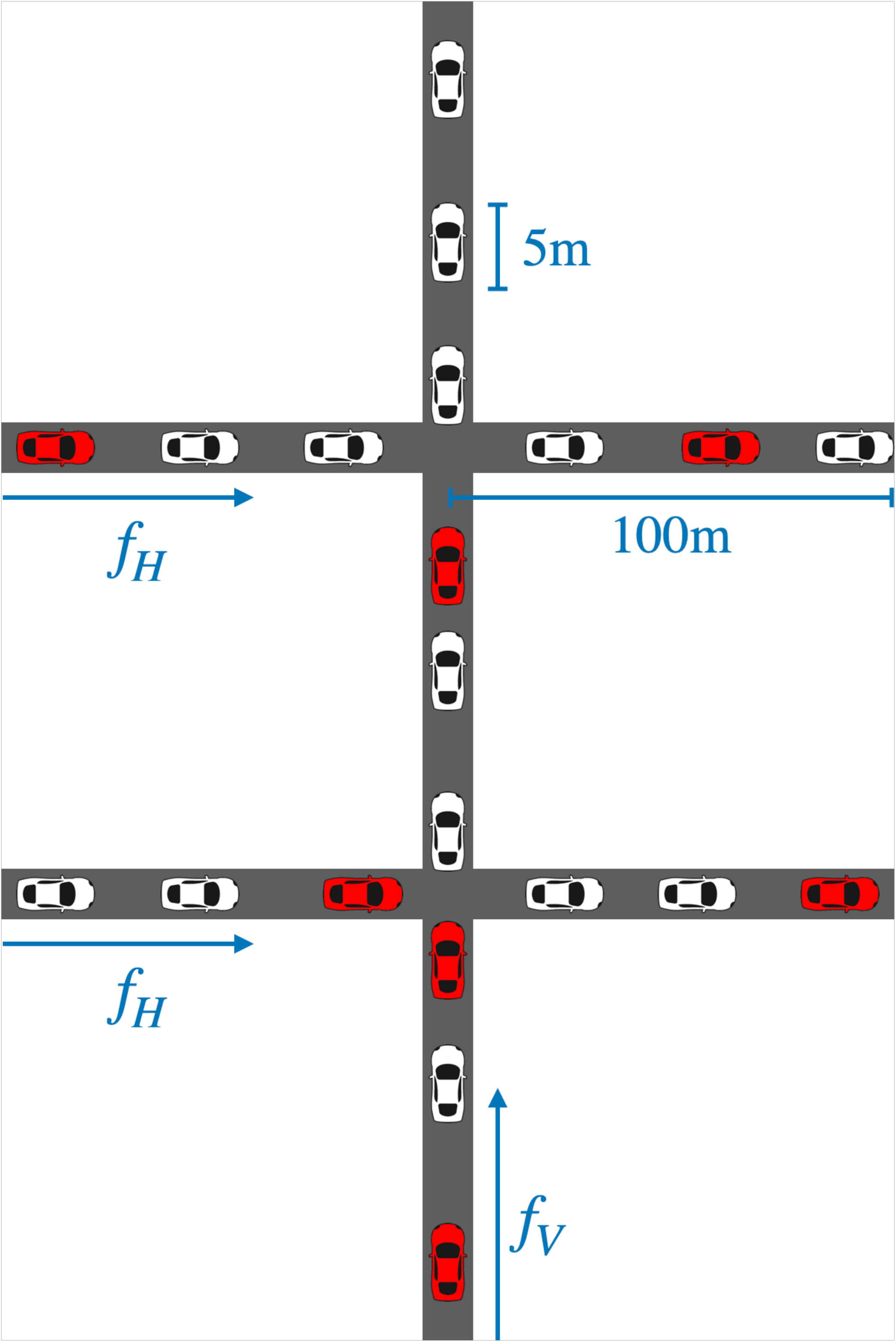}
    \hfil}
    \subfigure[Two-way 3x3]{\includegraphics[width=2in]{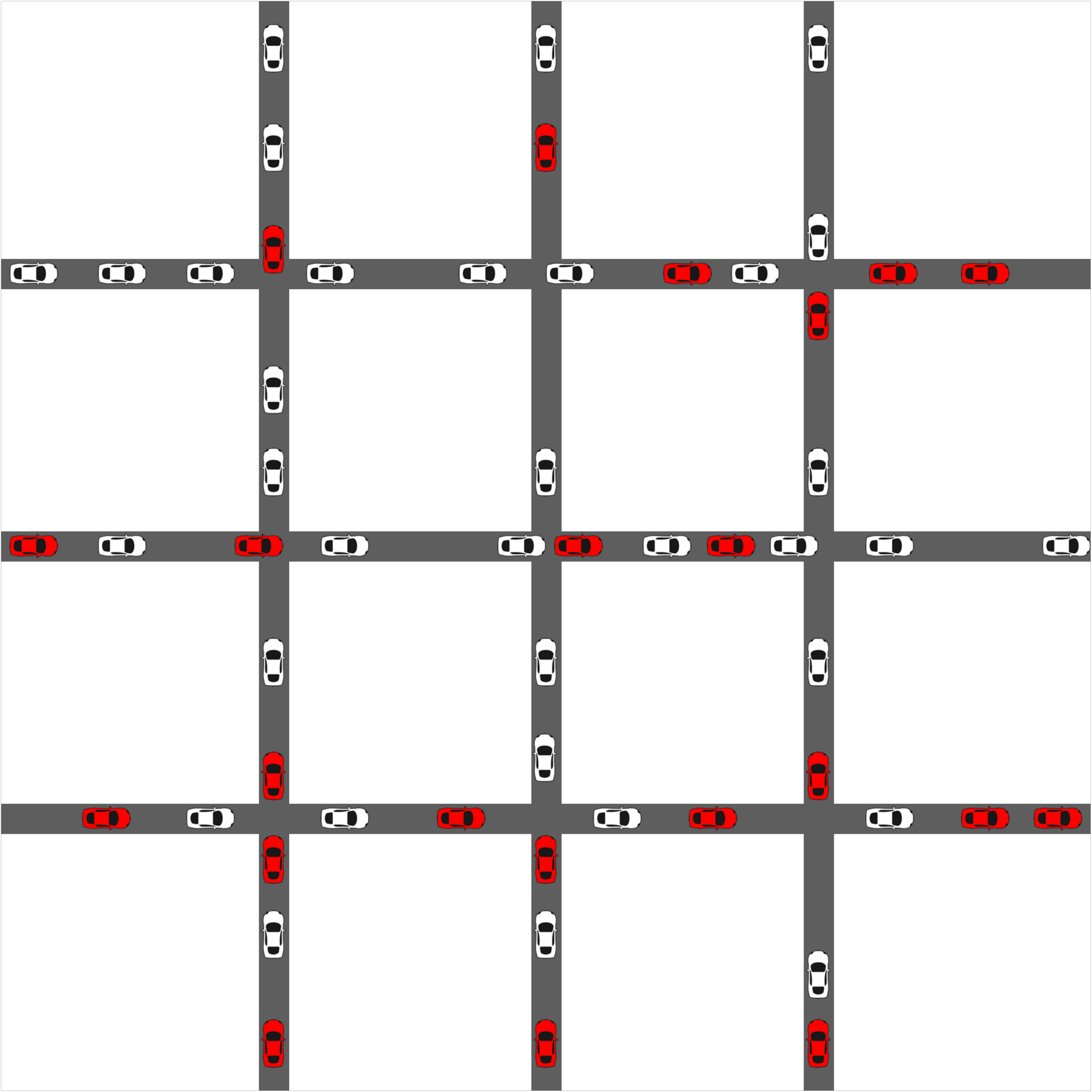}}
    \subfigure[Four-way 1x1]{\includegraphics[width=2in]{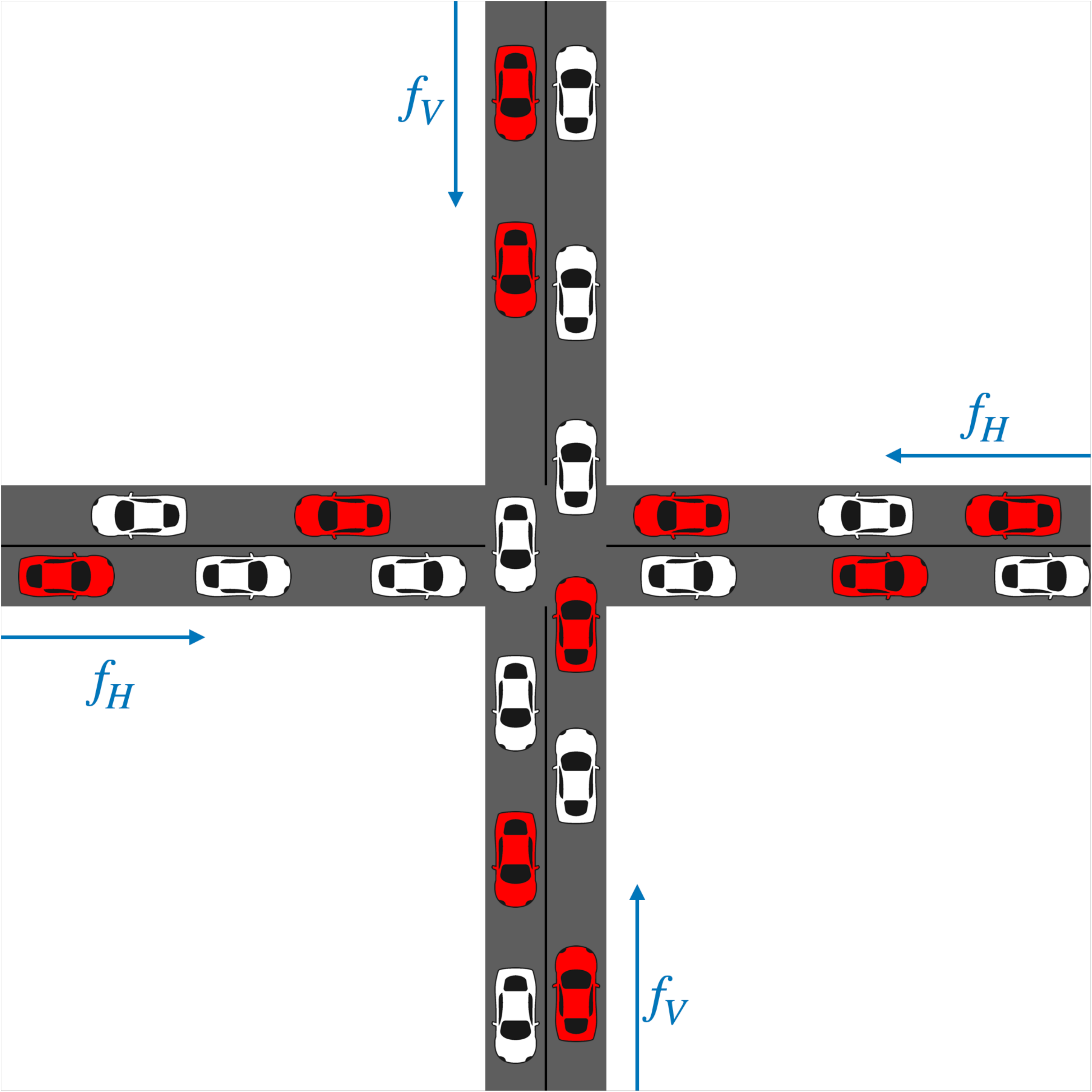}
    \hfil}
    \caption{\textbf{Traffic Networks}. For each network, every horizontal lane has inflow rate $f_H$ and every vertical lane has inflow rate $f_V$. Red vehicles represent AVs while white vehicles represent uncontrolled vehicles. Illustrations are not drawn to scale and vehicles are arbitrarily placed.}
    \label{fig:networks}
\end{figure*}

\section{Related Work}
Traditional adaptive traffic signal control methods such as SCOOT \cite{hunt1981scoot} and SCATS \cite{lowrie1990scats} are widely deployed in the real world and optimize the timing of traffic signals to minimize travel time in response to traffic volume. Recently, RL-based signal control methods demonstrate superior performance to traditional adaptive methods \cite{wei2019colight}. These works are orthogonal to our work, as mixed autonomy control may be combined with traffic signal control to optimize a given objective.

On the other hand, autonomous intersection management (AIM) methods generally control all vehicles near the intersection towards some objective \cite{miculescu2019polling}. Applying AIM to a mixed autonomy intersection, H-AIM \cite{sharon2017protocol} proposes a first-come-first-serve scheme to schedule AVs while guiding human vehicles with a fixed-time traffic signal. Our goal is fundamentally different: we would like to design a learning-based method to flexibly maximize any given objective.

The most similar works to ours are mixed autonomy works based on the Flow framework \cite{wu2017framework}, including \cite{wu2017emergent} and \cite{vinitsky2018lagrangian}. However, these works examine small scenarios with a constant number of control dimensions, and none address modular intersection networks, which are ubiquitous in real-world traffic networks. In contrast, we focus on mixed autonomy control of open intersection networks agnostic to the number of vehicles present at each step. Our work provide a key building block for scaling mixed autonomy to large traffic networks.

Our multi-agent RL method features the centralized training and decentralized execution paradigm, explored in multi-agent RL works such as PS-TRPO \cite{gupta2017cooperative} and MADDPG \cite{lowe2017multi}. In this paradigm, agents are jointly optimized at training time but execute independently afterwards.

\section{Preliminaries}

\subsection{Policy-based Model-free Reinforcement Learning}
\label{sec:policy_based_model_free_rl}
We frame a mixed autonomy traffic scenario as a finite-horizon discounted Markov Decision Process (MDP) defined by $(\cS, \cA, T, r, \rho_0, \gamma, H)$ consisting of state space $\cS$, action space $\cA$, stochastic transition function $T: \cS \times \cA \to \Delta(\cS)$, reward function $r: \cS\times \cA\times\cS \to \bbR$, initial state distribution $\rho_0: \cS \to \Delta(\cS)$, discount factor $\gamma \in [0, 1]$, and horizon $H \in \bbN_+$, where $\Delta(\cS)$ is the probability simplex over $\cS$.

Policy-based model-free RL algorithms define a policy $\pi_\theta: \cS \to \Delta(\cA)$ parameterized by $\theta$, which is often a neural network. In particular, the REINFORCE policy gradient algorithm \cite{williams1992simple} maximizes the expected discounted cumulative reward of policy $\pi_\theta$
\begin{equation}
    \max_{\theta} \bbE_{\substack{s_0\sim \rho_0,a_t\sim \pi_\theta(s_t)\\ s_{t+1}\sim T(s_t, a_t)}}\left[\sum_{t=0}^{H - 1} \gamma^t r(s_t, a_t, s_{t+1})\right]
\end{equation}
by sampling trajectories $(s_0, a_0\dots, s_{H-1}, a_{H-1}, s_H)$ and performing the following update
\begin{equation}
    \theta\leftarrow \theta + \alpha \sum_{t=0}^{H - 1} \nabla_\theta \left.\log \pi_\theta(s_t)\right\vert_{a_t}\sum_{t'=t}^{H - 1} \gamma^{t' - t} r(s_{t'}, a_{t'}, s_{t' + 1})
\end{equation}
where $\alpha$ is the learning rate.

\subsection{Mixed Autonomy Traffic}
The mixed autonomy traffic environment in microscopic simulation can be naturally modeled as a MDP. At each time step $t$, the state $s_t$ is composed of the positions, velocities, and other metadata of all vehicles in the road network. The action $a_t$ is the tuple of accelerations of all AVs in the road network at step $t$; the accelerations of the rest of the vehicles is given by a car-following model, specifically the Intelligent Driver Model (IDM) \cite{treiber2000congested} with Gaussian noise. The reward function $r$ is typically a function of throughput, fuel consumption, or safety in the traffic network. The stochastic transition function $T$ is not explicitly defined, but $s_{t+1} \sim T(s_t, a_t)$ can be sampled from the microscopic simulator, which applies the accelerations for all vehicles for a simulation step duration $\delta$. Like previous works in mixed autonomy, we use the SUMO microscopic simulator \cite{lopez2018microscopic}.

\section{Problem Setup: Simulated Traffic Scenarios}
We describe our method and baseline methods in the context of \textit{traffic scenarios}, each of which specifies a \textit{traffic network} and an \textit{inflow configuration}.
\begin{table}[htb]
\caption{\textbf{Inflow Configurations}. We consider the 16 inflow configurations marked with \checkmark. Each horizontal ($f_H$) or vertical ($f_V$) inflow rate has units of vehicle/hour/lane.}
\label{tab:inflow}
\begin{center}
\begin{tabular}{|c||c|c|c|c|c|}
\hline
\backslashbox{$f_H$}{$f_V$} & 400 &  550 &  700 &  850 & 1000 \\ \hline\hline
1000  &  \checkmark & \checkmark & \checkmark & \checkmark &      \\ \hline
 850  &  \checkmark & \checkmark & \checkmark & \checkmark & \checkmark \\ \hline
 700  &       &      & \checkmark & \checkmark & \checkmark \\ \hline
 550  &       &      &      & \checkmark & \checkmark \\ \hline
 400  &       &      &      & \checkmark & \checkmark \\ \hline \end{tabular}
\end{center}
\end{table}

\subsection{Traffic Networks}
As shown in Fig.~\ref{fig:networks}, our traffic networks are grids of single-lane two-way or four-way intersections with only through traffic and no turning traffic. Under a given inflow configuration, vehicles enter the traffic network via the inflow lanes at a specified inflow rate with equal spacing and leave the traffic network via outflow lanes. To populate the initial vehicles for $s_0 \sim \rho_0$, we run $H_0$ warmup steps where all AVs and uncontrolled vehicles follow the IDM.

\subsection{Inflow Configurations}
For every traffic network, we consider 16 inflow configurations shown in Table~\ref{tab:inflow}, which are chosen to exclude configurations with trivially low inflow or infeasibly high inflow. Each inflow configuration specifies a single inflow rate $f_H$ for every horizontal lane and another inflow rate $f_V$ for every vertical lane. If a vehicle is unable to inflow due to congestion, it is dropped from the simulation.

\subsection{Distribution of Autonomous Vehicles}
We test our mixed autonomy method under multiple penetration rates of AVs as a percentage of all vehicles. Unless otherwise stated, AVs are regularly-spaced on each lane of the traffic network (e.g. under 50\% AV penetration, every second vehicle is an AV).

\subsection{Traffic Objective}
Our main objective is to maximize throughput of the network, measured by outflow. In addition, to encourage safe intersection control and because higher collision implies lower outflow, our secondary objective seeks to minimize the collision rate. We believe that future works may easily adapt our method to other objectives involving combinations of throughput, fuel consumption, and safety.

\section{Reinforcement Learning for Mixed Autonomy}
Our method for mixed autonomy control is rooted in policy-based model-free RL as described in Section~\ref{sec:policy_based_model_free_rl}. Here we detail modifications specific to our method.

\begin{figure}[ht]
    \centering
    \includegraphics[width=0.8\columnwidth]{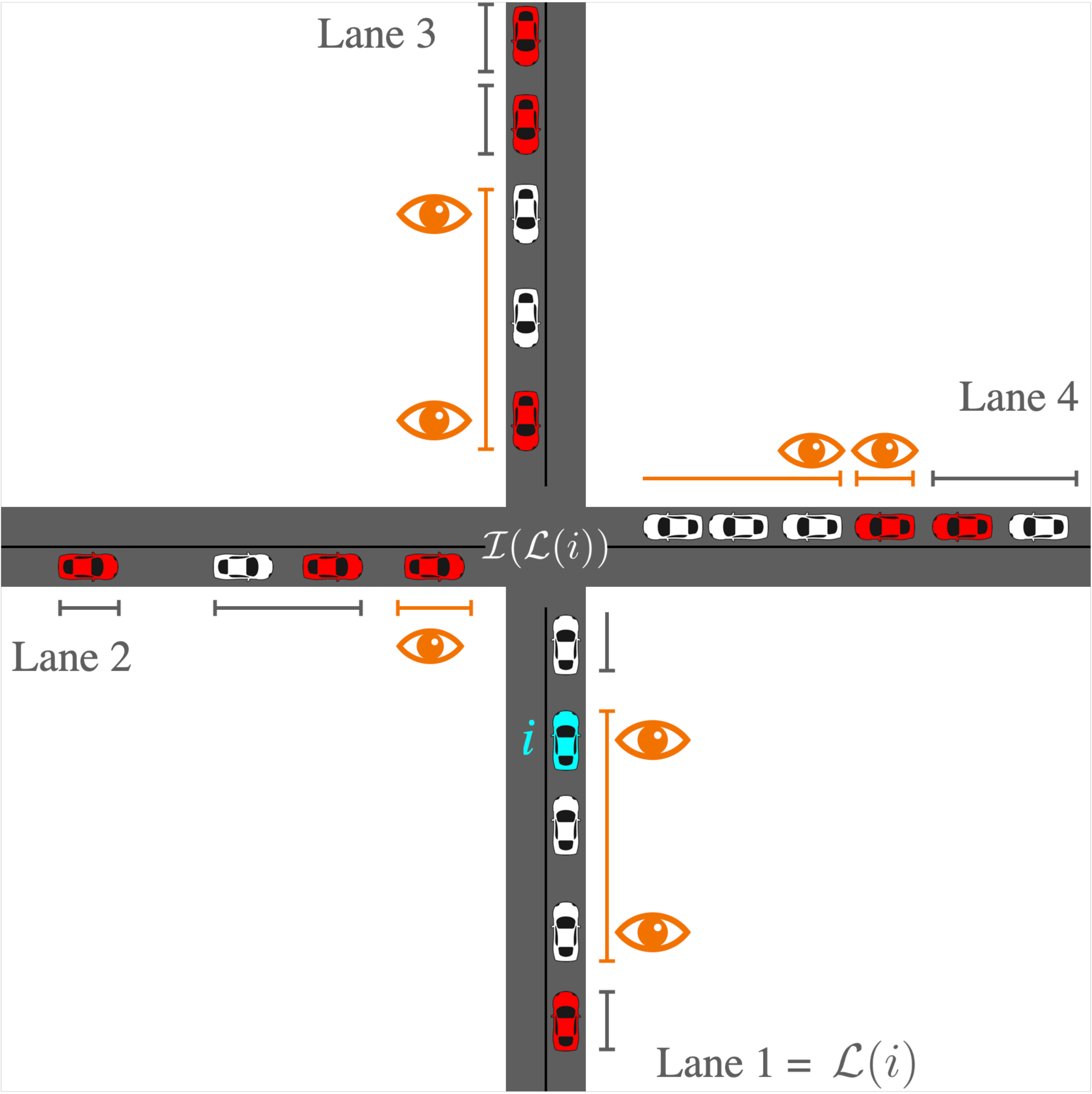}
    \caption{\textbf{Observation Function}. We illustrate the observation $z(s, i)$ of AV $i$, only showing vehicles approaching the intersection. The cyan vehicle represents the ego AV $i$, red vehicles represent other AVs, and white vehicles represent uncontrolled vehicles. Let $\mathcal L(i)$ be the lane of $i$ and $\mathcal{I}(\mathcal L(i))$ be the approaching intersection. For convenience, we number the lanes from 1 to 4, clockwise from lane $\mathcal L(i)$. Chains and half-chains are denoted by line-segments and half-segments, respectively; observed half-chains and chains are colored in orange; eye symbols are placed next to observed vehicles in observed chains. Notice that lanes 2 and 3 do not have a leading half-chain, so the observation is padded with default values.}
    \label{fig:obs}
\end{figure}
\subsection{Multi-agent Policy Decomposition}
To tractably optimize mixed autonomy traffic, we formulate the scenario as a decentralized partially observable MDP (Dec-POMDP) \cite{boutilier1996planning}. We decompose the policy $\pi_\theta: \cS \to \Delta(\cA)$ into per-AV policies $\pi^i_\theta: \cO^i \to \Delta(\cA^i)$ such that $\pi_\theta(s) = \prod_i \pi_\theta^i(o^i)$. $\cA^i$ is the acceleration space for AV $i \in \{1, \cdots, |\cA|\}$ and $\cO^i$ is the observation space for AV $i$ such that $o^i = z(s, i) \in \cO^i$ contains only a subset of the information of the state $s \in \cS$. We have $\cO^1 \times \cO^2 \times \dots \subseteq \cS$ and $\cA^1 \times \cA^2 \times \dots = \cA$. We define $z$ as the observation function which maps the observation $o^i$ for AV $i$ at state $s$. Our policy decomposition allows our policy to tolerate arbitrary number of AVs and arbitrarily large traffic scenarios; without decomposition, the combinatorial nature of $\cA$ poses an intractable problem to learning algorithms.

\subsection{Observation Function}
We define several helpful terminologies before defining our observation function. On any approaching lane $\mathcal L$ to intersection $\mathcal{I}(\mathcal L)$, we split $\mathcal L$'s vehicles into \textit{chains}, each led by an AV. We define the \textit{half-chain} to include any possible leading IDM vehicles in a lane.

As illustrated in Fig.~\ref{fig:obs}, the \textit{observed vehicles} for AV $i$ include the head and tail of $i$'s own \textit{chain} and, for each other lane in clockwise order, the tail of the \textit{half-chain} and the head and tail of the next \textit{chain}. If the half-chain or chain does not exist, we pad the observation with appropriate default values. We define our observation function $z(s, i)$ as the speed and distance to intersection for every \textit{observed vehicle}. While this definition introduces partial observability, it enables decentralized and local execution of the policy, as each per-vehicle policy only requires information near the approached intersection.

\subsection{Policy Architecture}
We define the per-vehicle policy $\pi_\theta^i$ as a neural network with three fully-connected layers. We share the policy parameter $\theta$ across all vehicles in the traffic network to share experiences between AVs \cite{gupta2017cooperative}. We use a discrete acceleration space $\cA^i = \{+c_\text{accel}, 0, -c_\text{decel}\}$, where the maximum acceleration $c_\text{accel}$ and maximum deceleration $c_\text{decel}$ are less than the maximums permitted to IDM vehicles.

\subsection{Reward Definition, Centering, and Normalization}
We define the unnormalized reward function
\begin{equation}
r'(s_t, a_t, s_{t+1}) = \lambda_o \text{outflow}(s_t, s_{t+1}) - \lambda_c \text{collisions}(s_t, s_{t+1})
\end{equation}
where $\text{outflow}(s_t, s_{t+1})$ is the global outflow during step $t$, i.e. the number of vehicles that exited the network during step $t$, and $\text{collisions}(s_t, s_{t+1})$ is the global number of collisions during step $t$. We acknowledge that this centralized reward results in a difficult objective to optimize in large traffic scenarios beyond the scope of this study, where reward decomposition into decentralized rewards may be necessary. We do not use any reward shaping in our experiments.

To reduce variance of policy gradients \cite{engstrom2019implementation}, we apply reward centering and normalization to get the reward function
\begin{equation}
r(s_t, a_t, s_{t+1}) = \frac{r'(s_t, a_t, s_{t+1}) - \hat \mu_{r'}}{\hat \sigma_{R'}}
\end{equation}
where $\hat \mu_{r'}$ is the running mean of $r'$ and $\hat \sigma_{R'}$ is the running standard deviation of the running cumulative reward, which is updated according to $R' \leftarrow \gamma R' + r'(s_t, a_t, s_{t+1})$.

\subsection{Multi-task Learning over Inflow Configurations}
\label{sec:multitask}
Given a traffic network and AV composition, we employ multi-task learning over all inflow configurations in Table~\ref{tab:inflow}. During training, we initialize one environment per inflow configuration. At each training step, our policy gradient algorithm receives trajectories from each environment and batches the gradient update due to these trajectories. Multi-task learning allows a single trained policy to generalize across multiple inflow configurations, avoiding the costs of training a separate policy for each inflow configuration.

\subsection{Transfer Learning and Finetuning}
In our experiments, we commonly transfer the trained policy from a source scenario (e.g. one with 50\% AVs) to initialize the policy for a target scenario (e.g. one with 15\% AVs) before finetuning the policy in the target scenario. We find that this practice tends to boost the policy performance on the target scenario, but we do not further analyze transfer learning's impact in this paper due to space constraint. We defer the full details of the transfer scheme and any intermediate model checkpoints to the code that we provide.

In particular, we apply zero-shot transfer from the smaller 2x1 scenario to the larger 3x3 target scenario, meaning that we evaluate the policy from the source scenario on the target scenario without finetuning.

\begin{figure*}[hbt]
    \centering
    \includegraphics[width=\textwidth]{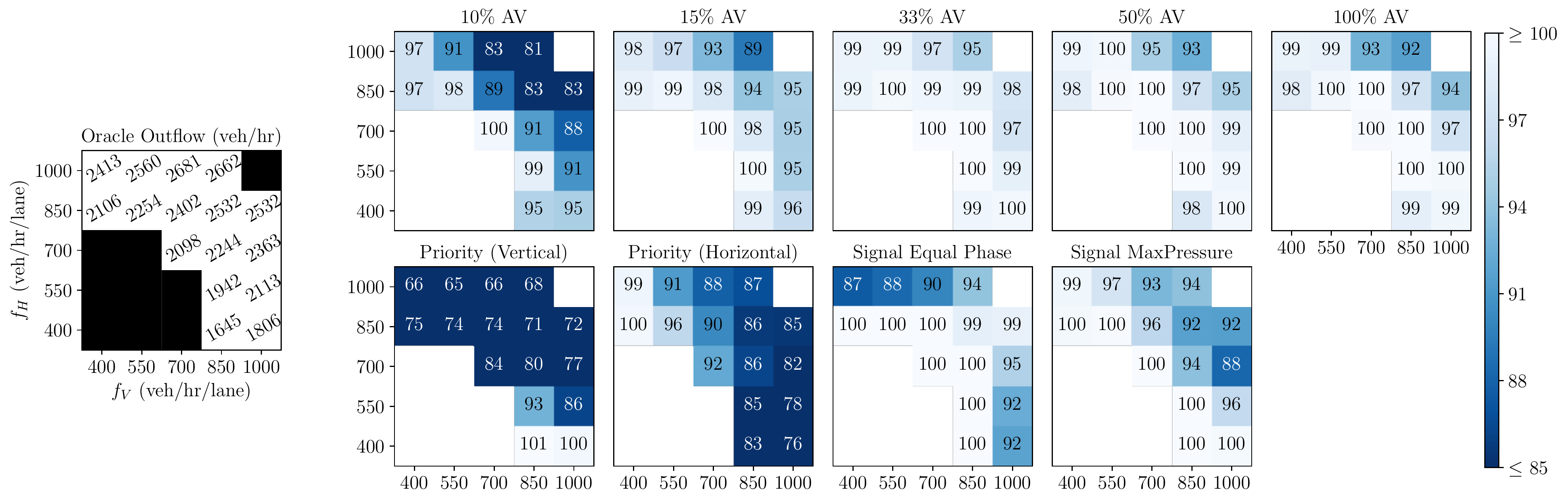}
    \caption{\textbf{Performance in Two-way 2x1 Scenarios}. The left table displays the Oracle method's hourly outflow, given the horizontal and vertical inflow rates $f_H$ and $f_V$. We consider these outflow rates to be optimal and express the hourly outflow of all other methods as a percentage of the Oracle outflow.}
    \label{fig:2x1}
\end{figure*}
\begin{figure}[hbt]
    \centering
    \includegraphics[width=\linewidth]{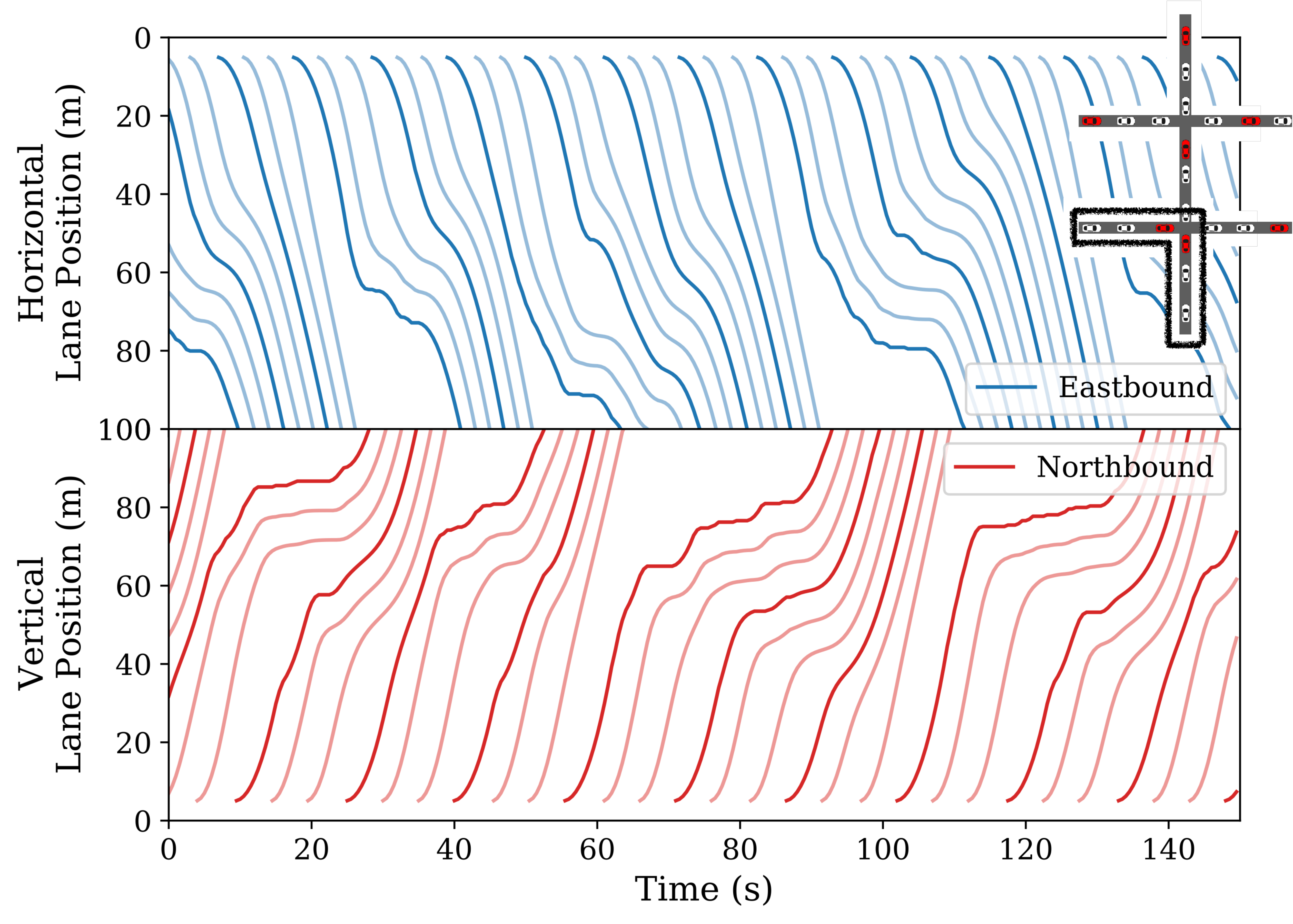}
    \caption{\textbf{Time-space Diagram of 33.3\% AV Policy in the Lower Intersection of Two-way 2x1 Scenario} with $f_H = 1000$ veh/hr/lane and $f_V = 700$ veh/hr/lane. The intersection is at position 100m along both the Northbound and Eastbound lanes. Lines with 100\% opacity denote AVs while lines with 50\% opacity denote uncontrolled vehicles. We see that the policy regulates the intersection similar to a traffic signal by letting groups of vehicles pass the intersection at once. The Eastbound lane, which has higher inflow rate, sees larger groups than the Northbound lane.}
    \label{fig:time_space_2x1}
\end{figure}

\section{Experimental Setup}

\subsection{Baseline Methods}
We compare numerical performances of our mixed autonomy method with signal-based or priority-based control methods, in which all vehicles are modeled with IDM.

Signal-based control methods offer anytime control at intersections, while mixed autonomy methods may only control traffic at intersections when AVs are present at the intersection. Therefore, under fixed inflow rates, we expect optimal signal-based methods to be upper bounds for optimal mixed autonomy methods under the throughput objective. For fairer comparison with mixed autonomy control, we set the yellow and all-red time of signals to 0s. We design the following signal-based baseline methods:
\begin{enumerate}
    \item \textit{Oracle}: for each scenario, we utilize a simple iterative hill climbing algorithm to empirically find the best horizontal and vertical signal phase length, which are shared for all traffic signals. By construction, this method achieves the optimal throughput for a traffic scenario with a known, fixed, and deterministic inflow distribution. Under other objectives or conditions, the Oracle method would likely not be optimal.
    \item \textit{Equal-phase}: we equip all traffic signals with equal phase lengths in both the horizontal and vertical directions. Given the traffic network, we empirically determine the phase length $\tau_\text{equal}$ which maximizes average performance across all inflow configurations.
    \item \textit{MaxPressure}: we implement adaptive traffic signals controlled by the MaxPressure algorithm, which is proven to maximize throughput in simplified intersections modeled by a queuing network \cite{varaiya2013max}. Given the traffic network, we empirically determine the minimum phase duration $\tau_\text{min}$ which maximizes average throughput across all inflow configurations.
\end{enumerate}
We use $\tau_\text{equal} = 25$s and $\tau_\text{min} = 4, 6, 12$s for Two-way 2x1, Two-way 3x3, and Four-way 1x1 respectively.

Priority-based baseline methods resemble placing a stop sign along the direction without priority. Therefore, we expect these baselines to serve as an illustrative lower bound in high inflow scenarios:
\begin{enumerate}
    \item \textit{Priority (Vertical)}: we assign priority to vertical lanes.
    \item \textit{Priority (Horizontal)}: we assign horizontal lanes with priority only in asymmetric scenarios.
\end{enumerate}

\subsection{Simulation Parameters}
For all scenarios in this study, we define lanes to be 100m long and vehicles to be 5m long. The speed limits are set to be 13m/s; uncontrolled vehicles may accelerate at up to 2.6m/s$^2$ and decelerate at up to 4.5m/s$^2$ while AVs' maximums are given by $c_\text{accel} = 1.5\text{m/s}^2$ and $c_\text{decel} = 3.5\text{m/s}^2$. For simulate each trajectory, we use a simulation step duration $\delta$ of 0.5s, $H_0 = 100$ warmup steps, and a training horizon $H = 2000$. To evaluate the equilibrium performance of trained policies or baseline methods, we execute the policy for trajectories of $500 + H$ steps then analyze only the last $H$ steps. All reported outflow or collision results are empirical means over 10 evaluation trajectories.

\subsection{Training Hyperparameters}
For training, we use outflow and collision coefficients $\lambda_o = 1$ and $\lambda_c = 5$. Our policy $\pi_\theta$ consists of two hidden layers of size 64. We use discount factor $\gamma=0.99$, learning rate $\alpha = 0.001$, and the RMSprop optimizer \cite{hinton2012neural}. In the paradigm of multi-task learning described in Section~\ref{sec:multitask}, each policy gradient update uses a batch of 640 total trajectories with multiple trajectories per inflow configuration. Starting from a random or pretrained initial policy, we perform up to 200 policy gradient updates to train or finetune the policy. We save a policy checkpoint every 5 gradient updates and select the checkpoint which achieves the best mean outflow across the training batch to evaluate.

\begin{figure*}[htb]
    \centering
    \includegraphics[width=\textwidth]{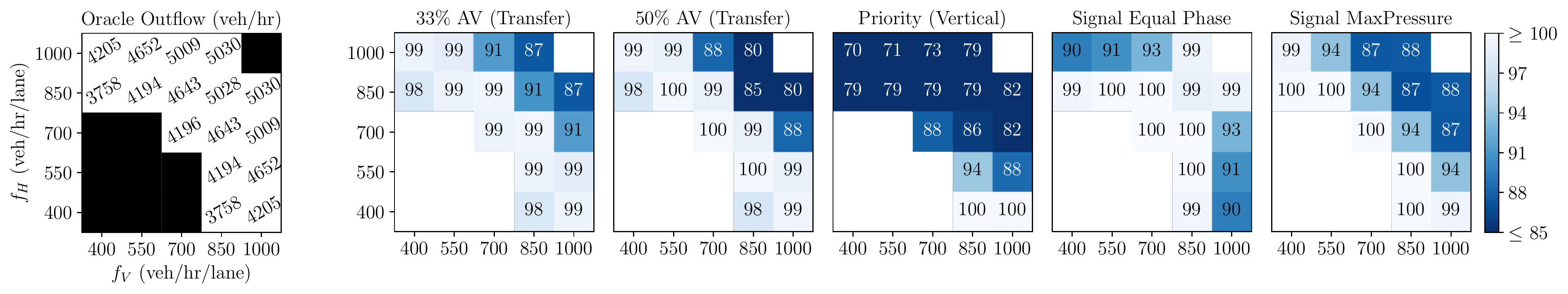}
    \caption{\textbf{Zero-shot Transfer Performance of Mixed autonomy Policies from Fig.~\ref{fig:2x1} to Two-way 3x3 Scenarios}.}
    \label{fig:3x3}
\end{figure*}

\begin{figure*}[htb]
    \centering
    \includegraphics[width=\textwidth]{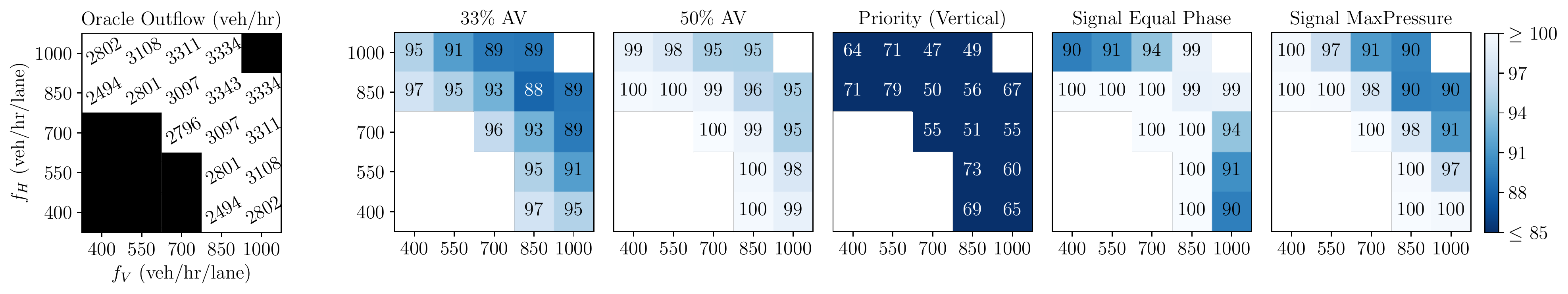}
    \caption{\textbf{Performance in Four-way 1x1 Scenarios}. We train and evaluate a 50\% AV policy in four-way 1x1 scenarios.}
    \label{fig:1x1_4way}
\end{figure*}

\section{Experimental Results}
\subsection{Two-way 2x1: Near-optimal Performance}
As seen in Fig.~\ref{fig:2x1}, our method produces near-optimal policies in two-way 2x1 traffic networks across the 16 inflow configurations. Each policy is trained and evaluated on the same AV penetration rate: 10\%, 15\%, 33.3\%, 50\%, or 100\%.

We observe that a 33.3\% penetration rate achieves the best throughput and that smaller (10\%, 15\%) and larger (50\%, 100\%) penetration rates of AVs see reduced performance; we believe that the former is due to underactuation and that the latter is due to increased optimization difficulty. Indeed, the highest inflow configurations typically see around 30 vehicles in the traffic network at a given time: this translates into a 30-AV coordination problem for the 100\% penetration rate scenario, which may be hard to optimize. On the other hand, the 10\% penetration rate scenario may only see 3 AVs in the network at a given time, which is likely insufficient for controlling intersection traffic.

Upon examining the learned policies, we find that the policies control AVs to implement a traffic-signal-like behavior in the \textit{unsignalized} intersection, as seen in Fig.~\ref{fig:time_space_2x1}. Note that this behavior emerges without any reward shaping, as we solely use an outflow reward and a collision penalty. This illustrates the potential of RL to discover advantageous behaviors without explicit human guidance.

\subsection{Two-way 3x3: Zero-shot Generalization}

In Fig.~\ref{fig:3x3}, we demonstrate zero-shot transfer and generalization of policies trained in two-way 2x1 scenarios to 3x3 grids of intersections. The optimality of both trained policies and baseline methods somewhat deteriorate as the scenario size increases, especially for high inflow configurations. Nevertheless, zero-shot transfer of mixed autonomy policies outperforms baseline methods.

\subsection{Four-way 1x1: Near-optimal Performance}
We apply the our method to four-way 1x1 scenarios and report the result for 33\% and 50\% AV in Fig.~\ref{fig:1x1_4way}. Notably, 50\% AV significantly outperforms 33\% AV in these scenarios. We hypothesize that these scenarios require more AVs to coordinate between more lanes.

\section{Discussion}
Our work shows that multi-agent reinforcement learning learns near-optimal policies in mixed autonomy traffic under regular conditions and exhibit generalization across inflow rates and network size. We view our study as a first exploration of traffic control in mixed autonomy scenarios, and we discuss several followup directions here.

First, we note that our study focuses on outflow as a metric of throughput, while another commonly used metric for intersection control is the total trip time. We do not use the total trip time here because congested vehicles may build up linearly in high inflow scenarios, and therefore the total trip time may increase linearly with the horizon $H$, while the outflow does not depend on $H$. In addition, we do not consider fuel consumption for this study, and the choice of a discrete action space $\{+c_\text{accel}, 0, -c_\text{decel}\}$ may increase fuel consumption. We believe that future work may consider other metrics of optimization and address fuel consumption by training directly on a continuous action space or applying imitation learning and/or policy distillation to obtain a continuous policy from a discrete policy \cite{rusu2016policy}.

Second, we acknowledge that our method would likely induce a distribution shift in human driver behavior if deployed in the real world. This is a fundamental issue in mixed autonomy and other adaptive control methods, though we speculate that the distribution shifts induced by AVs may be larger as humans drivers may be unaccustomed to being shepherded by AVs, especially in unsignalized intersections.

Finally, we reiterate that our work in mixed autonomy control of intersections is an important stepping stone towards adaptive integration of mixed autonomy and traffic signal control mechanisms. Such integration may facilitate joint optimization of throughput, fuel consumption, and safety objectives beyond the capabilities of either mixed autonomy or traffic signals alone, while traffic signal control may help alleviate distribution shift due to mixed autonomy control.


\section*{ACKNOWLEDGMENT}
The authors acknowledge MIT SuperCloud and Lincoln Laboratory Supercomputing Center for providing computational resources that have contributed to the research results in this paper. Zhongxia Yan was partially supported by the Department of Transportation Dwight David Eisenhower Transportation Fellowship Program (DDETFP).

\addtolength{\textheight}{-20cm} 

\bibliographystyle{IEEEtran}
\bibliography{IEEEabrv,reference}

\end{document}